%% file: acl_latex.tex
\pdfoutput=1
\documentclass[11pt]{article}
\usepackage[]{EACL2023}

\usepackage{times}
\usepackage{latexsym}
\usepackage[T1]{fontenc}
\usepackage[utf8]{inputenc}
\usepackage{microtype}
\usepackage{url}
\usepackage{amsmath,graphicx}
\usepackage{amssymb,amsfonts}
\usepackage{textcomp,booktabs}
\usepackage{xcolor}
\usepackage{mathtools}
\usepackage{subfigure}
\usepackage{multirow}
\usepackage[normalem]{ulem}
\useunder{\uline}{\ul}{}

\input{math_commands.tex}

\DeclareMathOperator*{\encoder}{Trans-Enc}
\DeclareMathOperator*{\decoder}{Trans-Dec}
\DeclareMathOperator*{\lmhead}{LM-Head}
\DeclareMathOperator*{\mlp}{MLP}
\DeclareMathOperator*{\relu}{ReLU}

\title{Multimodal Event Transformer for Image-guided Story Ending Generation}

\author{Yucheng Zhou, Guodong Long \\
         Australian AI Institute, School of Computer Science, FEIT, University of Technology Sydney \\
         {\tt yucheng.zhou-1@student.uts.edu.au, guodong.long@uts.edu.au}\\
         }

\begin{document}
\maketitle
\begin{abstract}
Image-guided story ending generation (IgSEG) is to generate a story ending based on given story plots and ending image. Existing methods focus on cross-modal feature fusion but overlook reasoning and mining implicit information from story plots and ending image. To tackle this drawback, we propose a multimodal event transformer, an event-based reasoning framework for IgSEG. Specifically, we construct visual and semantic event graphs from story plots and ending image, and leverage event-based reasoning to reason and mine implicit information in a single modality. Next, we connect visual and semantic event graphs and utilize cross-modal fusion to integrate different-modality features. In addition, we propose a multimodal injector to adaptive pass essential information to decoder. Besides, we present an incoherence detection to enhance the understanding context of a story plot and the robustness of graph modeling for our model. Experimental results show that our method achieves state-of-the-art performance for the image-guided story ending generation.
\end{abstract}

\section{Introduction}
Story ending generation \cite{Guan19Story} aims to generate a reasonable ending for a given story plot. It requires deep models to integrate powerful language understanding capability, which is crucial for artificial intelligence. Many efforts \cite{Wang19TCVAE,Guan19Story,Yao19Plan,Guan20Knowledge} have been proposed and achieved promising results since neural models designed for comprehending natural language allow them to understand story plots and reason reasonable story endings. With the advance of automatic story generation, it has attracted outstanding attention in multimodality research \cite{Jung20Hide, Yu21Transitional, Chen21Commonsense}.

\begin{figure}[t]
  \centering
  \includegraphics[width=\linewidth]{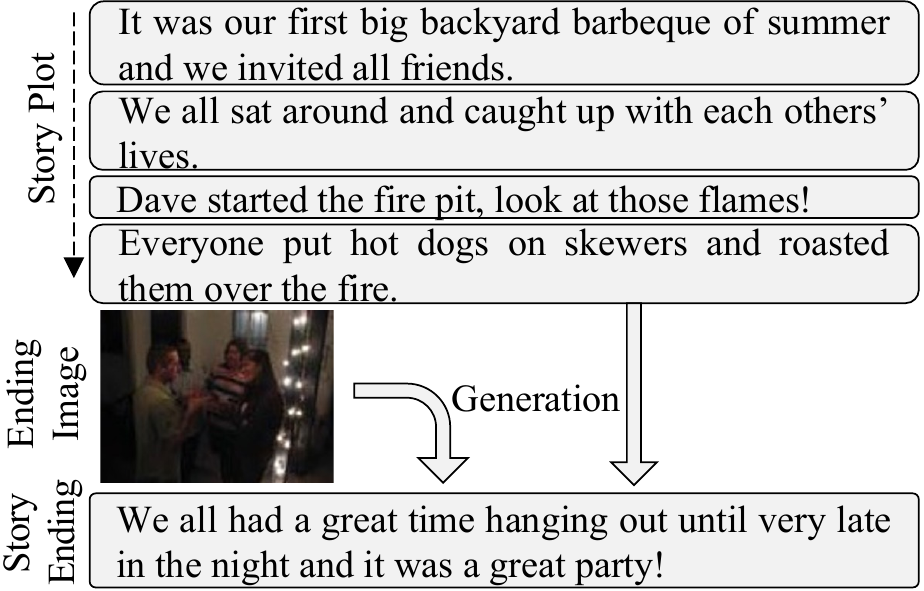}
  \caption{Given a multi-sentence story plot and an ending image, the image-guided story ending generation aims to generate a story ending related to the image.}
  \label{fig:intro}
\end{figure}

However, since story plots and story ending usually correspond to different content, the context with information bottleneck is not enough to deduce an informative story ending, i.e., generated endings tend to be inane and generic. To address this issue, \citet{Huang21IgSEG} propose an image-guided story ending generation (IgSEG) task that combines story plots and ending image to generate a coherent, specific and informative story ending. IgSEG demands not only introducing information from the ending image to story plots for story ending generation but also reasoning and mining implicit information from story plots and ending image, respectively. As shown in Figure \ref{fig:intro}, for story plots, ``party'' can be inferred from ``big backyard barbeque'' and ``invited all friends'', and ``all friends'', ``all sat around'' and ``caught up with'' can deduce ``had a great time''. For the ending image, ``dim indoor'' and ``bright lights'' can infer ``very late in the night''.

Existing methods \cite{Huang21IgSEG,Xue22MMT} focus on cross-modal feature fusion but overlook reasoning and mining implicit information from story plots and ending images. Nonetheless, to effectively conduct cross-modal feature fusion, it is necessary to reason and mine more implicit information from single-modality data. An event is a fine-grained semantic unit, which refers to a text span composed of a predicate and its arguments \cite{Zhang20ASER}. Recently, event-centric reasoning displays excellent capability for context understanding and subsequent event prediction \cite{Zhou22ClarET}. In this work, we propose a multimodal event transformer (MET) to mine implicit information to improve cross-modal fusion. For story plots, we leverage semantic role labeling (SRL) parser \cite{He17srl} to extract events from story plots and then construct them into a semantic event graph. For an ending image, we utilize scene graph parser \cite{Zellers18Motifs} to capture visual concepts and their relation to construct visual event graphs. Since edges contain relationships between nodes in visual and semantic event graphs, we employ relational graph convolutional networks (RGCN) \cite{Schlichtkrull18Modeling} to encode event graphs to infer implicit information.

For cross-modal feature fusion, most recent works \cite{Huang21IgSEG,Xue22MMT} adopt attention-based neural network models to implicitly integrate multi-modal features. However, due to the complexity of cross-modal features and the existence of dependency between single-modal features, it is often difficult for these models to complement cross-modal features. To tackle the issue, we propose cross-modal fusion to integrate different-modality features. Specifically, we merge visual and semantic event graphs and use RGCN to fuse cross-modal features for feature complement. 

Moreover, since features from different modalities suffer from domain inconsistency, previous methods \cite{Huang21IgSEG,Xue22MMT} directly concatenate them and pass them to the decoder, which is not a crafted manner. To appropriately combine features from different modalities, we design a multimodal injector to integrate relevant features into the decoder. In addition, we propose an incoherence detection to enhance the context understanding for a story plot and the robustness of graph modeling for our model.

In experiments, we conduct extensive evaluations on two datasets (i.e., VIST-E \cite{Huang21IgSEG} and LSMDC-E \cite{Xue22MMT}). Experimental results show that our method outperforms strong competitors and achieves state-of-the-art performance. In addition, we conduct further analysis to demonstrate the effectiveness of our method. Lastly, we compare the performance of our method and other methods through human evaluation.

\section{Related Work}
\subsection{Story Ending Generation}
Story ending generation aims to generate a story ending for given story plots, and it is one of the important tasks in natural language generation. Many efforts have been invested in story ending generation \cite{Wang19TCVAE,Guan19Story,Yao19Plan,Guan20Knowledge}. To make the generated story ending more reasonable, \citet{Guan19Story} propose a model encapsulating a multi-source attention mechanism, which can utilize context clues and understand commonsense knowledge. To ensure the coherence in generated story endings, \citet{Wang19TCVAE} propose a transformer-based conditional autoencoder, which can capture contextual clues in story splot. To improve long-range coherence in generated stories, \citet{Guan20Knowledge} pre-train model on external commonsense knowledge bases for the story ending generation. \citet{Zhou22ClarET} propose a correlation-aware context-to-event pre-trained transformer, which applies to a wide range of event-centric reasoning and generation scenarios, including story ending generation. Beyond the limit of single-modal information, \citet{Huang21IgSEG} introduce visual information to enrich the generation of story endings with more coherent, specific, and informative. To improve cross-modal feature fusion, \citet{Xue22MMT} propose a multimodal memory transformer, which fuses contextual and visual information to capture the multimodal dependency effectively.

\subsection{Visual Storytelling}
Visual storytelling task is proposed by \citet{Huang16Visual}, which aims to generate a story based on a given image stream. \citet{Wang18No} present an adversarial reward learning framework to learn an implicit reward function from human demonstrations. To inject imaginary concepts that do not appear in the images, some works \cite{Yang19Knowledgeable,Chen21Commonsense,Xu21Imagine} propose building scene graphs and injecting external knowledge into model to reason the relationship between visual concepts. \citet{Qi21Latent} propose a latent memory-augmented graph transformer to exploit the semantic relationships among image regions and attentively aggregate critical visual features based on the parsed scene graphs.

\begin{figure*}[t]
  \centering
  \includegraphics[width=\linewidth]{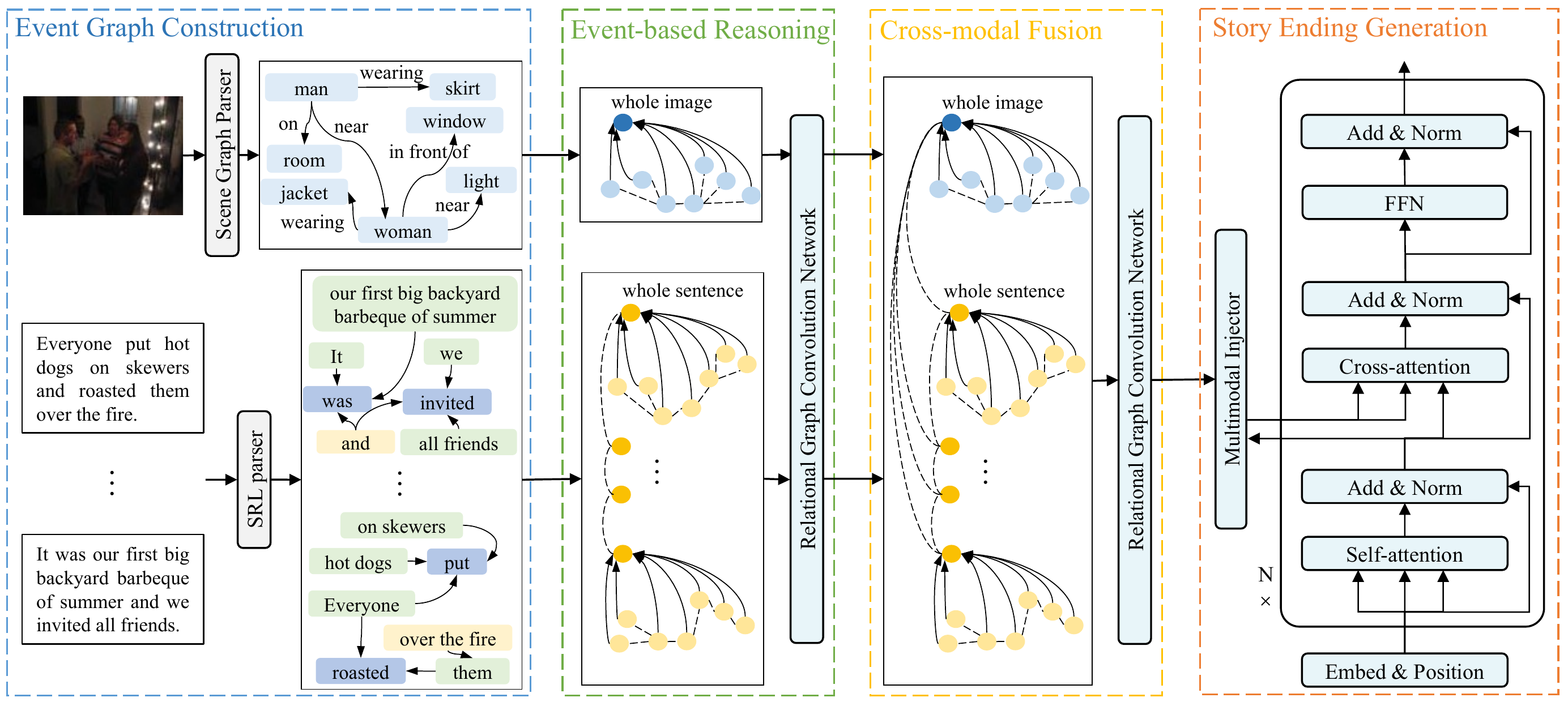}
  \caption{An overview of our model. Grey rounded rectangles denote fixed model. Blue rounded rectangles denote parameters that will be optimized.}
  \label{fig:model}
\end{figure*}

\subsection{Event-centric Reasoning}
Events always play an essential role in a story because a story is composed of multiple events and implies the relationship between the events. An event is a text span composed of a predicate and its arguments \cite{Zhang20ASER}. Multiple events include relations between events that conform to human commonsense \cite{Zhou22EventBERT}. Some works use plot events for story generation, which is generating a prompt and then transforming it into a text \cite{Ammanabrolu20Story,Fan19Strategies}. To generate a more coherent and specific ending, understanding events in story plots and their relationship can obtain informative context, which is a crucial step for story ending generation.

\section{Method}
This section will elaborate on our method for image-guided story ending generation, including event graph construction, event-based reasoning, cross-modal fusion, multimodal injector and story ending generation. The details of our method are shown in Figure~\ref{fig:model}. Lastly, details about objectives and training are elaborated.

\subsection{Event Graph Construction}
\paragraph{Semantic Event Graph.} 
The story plot contains multiple events which are correlated with each other. The definition of an event is a text span composed of a predicate and its arguments \cite{Zhang20ASER}. The event-centric reasoning shows excellent capability for context understanding and subsequent event prediction \cite{Zhou22ClarET}. To effectively reason and mine more implicit information from story plots, we use semantic role labeling (SRL) to parse the story and extract events from parsing results, as shown in Figure~\ref{fig:model}. Specifically, Given story plots $\gS=\{\mS_1, \mS_2, \mS_3, \mS_4\}$, we construct semantic event graphs $\gG^s_i=(\gV^s_i, \gE^s_i)$ by SRL. $\gE^s_i$ consists of two vectors, one for the positive direction and one for the opposite direction, and $\gV^s_i = \{\vs_0^i, \vs_1^i, \vs_2^i, \cdots, \vs_n^i\}$. To obtain features of each node, we use a pre-trained transformer encoder to obtain token representations in sentence $\mS_i$.
\begin{align}
\mT_i = \encoder(\mS_i), \mT_i \in \{\vt_i^1, \vt_i^2, \cdots, \vt_i^g\}
\end{align}
where $\vt_i^g$ denotes token representation, and $g$ is length of sentence $\mS_i$.
Next, we conduct a mean pooling operation for tokens presentations based on SRL parsing result $\hat \mS_i$ to get presentation $\hat \vs_j^i$ for each node. In addition, we take pooling for all token presentations of sentence $\mS_i$ to obtain a presentation of sentence node $\hat \vs_0^i$. Each node $\hat \vs_j^i$ in sentence $\mS_i$ is connected to the sentence node. To preserve the relationship between sequences, we connect sentence nodes in the order of the sequence.

\paragraph{Visual Event Graph.} 
For ending images, previous works \cite{Huang21IgSEG,Xue22MMT} use pre-trained convolutional neural networks (CNN) to extract feature maps directly. We construct visual event graphs to reason and mine more implicit information from ending images. Scene graphs have been used for many tasks to produce structured graph representations of visual scenes \cite{Zellers18Motifs}. Inspired by the success of these tasks, we parse the ending image $\mI$ to a scene graph via the scene graph parser. A scene graph can be denoted as a tuple $\gG^I=\{\gV^I, \gE^I\}$, where $\gV^I=\{\vv_0, \vv_1, \vv_2, \cdots, \vv_k\}$ is a set of $k$ detected objects. $\vv_0$ denotes a representation of the whole image, and other $\vv_i$ is a region representation of detected object. $\gE^I=\{e_1, e_2, \cdots, e_m\}$ is a set of directed edges and each edge $e_i$ refers to a triplet $(\vv_i, \vr_{i,j}, \vv_j)$, which includes two directional edges from $\vv_i$ to $\vr_{i,j}$ and from $\vr_{i,j}$ to $\vv_j$. Specifically, the construction of the scene graph can be divided into two parts: one is object detection, and the other is visual relation detection.

For object detection, we leverage a well-trained object detector, Faster-RCNN \cite{Ren17Faster} with a ResNet-152 \cite{He16Deep} backbone, to classify and encode objects in the ending image $\mI$. The outputs of detector include a set of region representations $\gV^I=\{\vv_1, \vv_2, \cdots, \vv_k\}$ and object categories $\gO=\{o_1, o_2, \cdots, o_k\}$. For visual relation detection, we leverage MOTIFS \cite{Zellers18Motifs} as our relation detector to classify the relationship between objects. We train the relation detector on Visual Genome dataset \cite{Krishna17Visual}. The output of relation detector is a set of relation $\gE^I=\{e_1, e_2, \cdots, e_m\}$, where $e_i$ refers to a triplet $(\vv_i, \vr_{i,j}, \vv_j)$. Lastly, we obtain the scene graph $\gG^I=\{\gV^I, \gE^I\}$ of ending image by combining the results of object detection and relationship detection.

\subsection{Event-based Reasoning}
We perform graph-structure reasoning over semantic and visual event graphs to effectively reason and mine more implicit information from story plots and ending images. Since event graphs have multiple relations between nodes (e.g., relations between visual objects, relations between predicates and arguments, etc.), we select relational graph convolutional networks (RGCN), which can pass different messages along different relations. Specifically, for each layer $l$ in $L$-layer RGCN, the node representation $\vw^l_i$ is updated as follows:
\begin{align}
\vw^{l+1}_i\!\!=\!\!\relu\!\!\Big(\sum_{r \in \gR} \sum_{j \in \gN_{r}(i)} \frac{1}{\left|\gN_{r}(i)\right|} \mW_{r} \cdot \vw_{j}^{l}\Big) \label{equ:rgcn}
\end{align}
where $\gR$ denote a set of all edges types, and $\gN_{r}(i)$ is the neighborhood of node $i$ under relation $r$. To reason and mine more implicit information in single-modality, we conduct event-based reasoning on semantic and visual event graphs, respectively.

\subsection{Cross-modal Fusion}
We propose cross-modal fusion for visual and semantic event graphs to integrate information from story plots and ending images. We adopt a layer normalization for node features to reduce the cross-modal gap between visual and semantic graphs. For cross-modal feature fusion, previous works \cite{Huang21IgSEG,Xue22MMT} adopt attention-based neural network models to implicitly integrate multi-modal features. However, these models neglect the dependency between single-modal features. Therefore, we maintain graph structure for visual and semantic features and connect nodes that represent whole image and sentences, as shown in Figure \ref{fig:model}. Moreover, we utilize RGCN as Eq.\ref{equ:rgcn} to integrate cross-modal features in event graph, and outputs denote as $\bar \gV^s_i = \{\bar \vs_0^i, \bar \vs_1^i, \bar \vs_2^i, \cdots, \bar \vs_n^i\}$ and $\bar\gV^I=\{\bar\vv_0, \bar\vv_1, \bar\vv_2, \cdots, \bar\vv_k\}$.

\subsection{Multimodal Injector}
\begin{figure}[t]
  \centering
  \includegraphics[width=0.9\linewidth]{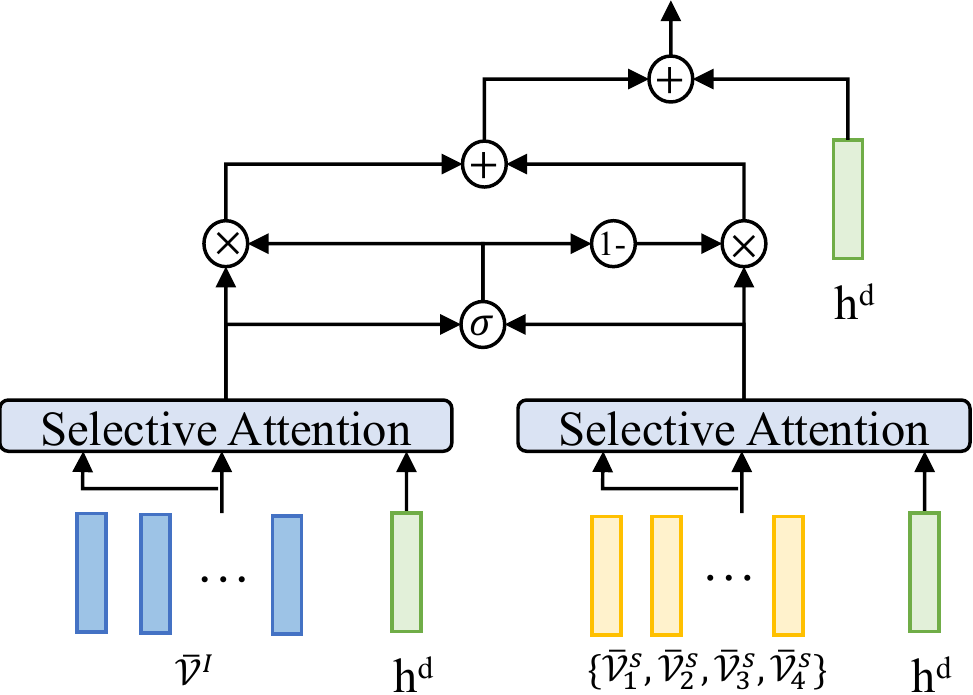}
  \caption{Details of the multimodal injector.}
  \label{fig:module}
\end{figure}
To integrate different modal sources, we propose a multimodal injector, which adaptly extracts key information from different modal features and integrates them appropriately. As shown in Figure \ref{fig:module}, inputs of multimodal injector include a hidden state $\vh^d$ from the decoder, visual features $\bar\gV^I$ and semantic features $\bar \gV^s_i$. Specifically, we first use selective attention for key information extraction, i.e., 
\begin{align}
\vh^u_{attn} = \softmax\left(\frac{Q K^{\mathrm{T}}}{\sqrt{d_{k}}}\right) V, u \in \{I,S\}
\end{align}
where Q is $\vh^d$ from decoder; K and V are visual features $\bar\gV^I$ or semantic features $\bar \gV^s_i$; and $d_{k}$ is the same as the dimension of $\vh^d$. Then, the gate $\lambda \in$ [0, 1] and the fused output are defined as:
\begin{align}
\lambda=\sigmoid\left(\mU \vh^I_{attn}+\mV \vh^S_{attn}\right)
\end{align}
where $\mU$ and $\mV$ are trainable weights. $\lambda$ controls how much visual information is attended.
\begin{align}
\hat \vh^d=\lambda \cdot \vh^I_{attn} + (1-\lambda) \cdot \vh^S_{attn} + \vh^d
\end{align}
where the fusion vector $\hat \vh^d$ is fed into the decoder.

\subsection{Story Ending Generation}
Recently, Transformer \cite{Vaswani17Attention} shows its powerful ability to generate natural language \cite{radford2019language}. For story ending generation, we use a Transformer decoder as the decoder for our model. Specifically, the decoder input includes a segment of the generated story ending $\bar{\mC}$ and fusion vector $\hat \vh^d$ from the multimodal injector. The purpose of the decoder is to predict a probability distribution of the next word of the segment $\bar{\mC}$, i.e.,
\begin{align}
\notag \vh_{i} &= \decoder(\hat \vh^d, \bar{\mC}) \in \R^d \\
    &\text{where}~\bar{\mC} = [c_1, \dots, c_{i-1}] \\ 
\vp_{i} &= \lmhead(\vh_{i}) \in \R^V
\end{align}
where $\vh_i$ refers to the hidden representation in $i$-th step; $V$ denotes token vocabulary and $\vp_{i}$ refers to a probability distribution over $\gV$; $d$ in $\hat \vh^d$ denotes the current number of layer. Lastly, the story ending generation objective is defined as a maximum likelihood estimation. The loss function is defined as:
\begin{align}
  \gL^{(gen)} = - \dfrac{1}{|N|}\sum\nolimits_{i=1}^{N}\log \vp_{i}(c_i), \label{equ:loss_cap}
\end{align}
where $\vp_{i}(c_i)$ denotes fetching the probability of the $i$-th step gold token $c_i\in\mC$ from $\vp_{i}$. $\mC$ refers to the gold caption, and $N$ is its length.

\subsection{Incoherence Detection}
To enhance the understanding context of a story plot and robustness of graph modeling for our model, we introduce a training objective: incoherence detection. We set a 10\% probability to replace a whole sentence node in semantic event graph randomly. In the objective, the final step output $\vh_n$ of the decoder is passed into a MLP to classify whether each whole sentence node is changed, i.e.,
\begin{align}
  \vp^{clf} &= \sigma(\mlp(\vh_n)) \in \R^4
\end{align}
where $\sigma$ denotes a sigmoid function. The loss function is defined as:
\begin{align}
  \notag \gL^{(clf)} &=-\frac{1}{4} \sum_{i=1}^{4} y_{i} \cdot \log (\vp^{clf}_i)\\
  &+(1-y_{i}) \cdot \log(1-\vp^{clf}_i)
\end{align}

\subsection{Training}
In model training, we set a trade-off parameter $\alpha$ for two losses $\gL^{(gen)}$ and $\gL^{(clf)}$. The total loss function of our model is definite as follows:
\begin{align}
  \gL = \gL^{(gen)} + \alpha \times \gL^{(clf)} \label{equ:loss}
\end{align}

\section{Experiment}
\subsection{Dataset and Evaluation Metric}
\paragraph{VIST-Ending.}
We compare our model and other state-of-the-art methods on the VIST-Ending (VIST-E) dataset \cite{Huang21IgSEG}. The dataset is built over VIST dataset \cite{Huang16Visual}. The VIST-E dataset comprises 39,920 samples for training, 4,963 samples for validation and 5,030 samples for testing. In experiments, we follow the data split in \cite{Huang21IgSEG}.
\paragraph{LSMDC-Ending.}
LSMDC-Ending (LSMDC-E) \cite{Xue22MMT} contains 20,151 training samples, 1,477 validation samples and 2,005 test samples, which are collected from LSMDC 2021 \cite{Rohrbach17Movie}.
\paragraph{Visual Genome.}
We use the Visual Genome (VG) dataset to train a visual relationship detector. 
The dataset includes 108,077 images annotated with scene graphs, and we follow the setting in \cite{Xu17Scene}, which contains 150 object classes and 50 relation classes.
\paragraph{Evaluation Metric.}
As follow \citet{Xue22MMT}, we utilize the same metrics to report evaluation results, and the evaluation code is open-source$\footnote{https://github.com/tylin/coco-caption}$. The evaluation metrics include: BLEU \cite{KingmaB14}, METEOR \cite{Banerjee05METEOR}, CIDEr \cite{vedantam2015cider}, ROUGE-L \cite{lin2004rouge} and Result Sum (rSUM) \cite{Xue22MMT}.

\begin{table*}[t]
\centering
\begin{tabular}{lcccccccc}
\toprule
\textbf{Method} & \textbf{B@1}   & \textbf{B@2}  & \textbf{B@3}  & \textbf{B@4}  & \textbf{M}     & \textbf{R-L}   & \textbf{C}     & \textbf{rSUM}   \\\midrule
Seq2Seq \cite{Luong15Effective}        & 13.96          & 5.57          & 2.94          & 1.69          & 4.54           & 16.84          & 12.04          & 57.58           \\
Transformer \cite{Vaswani17Attention}    & 17.18          & 6.29          & 3.07          & 2.01          & 6.91           & 18.23          & 12.75          & 66.44           \\
IE+MSA \cite{Guan19Story}         & 19.15          & 5.74          & 2.73          & 1.63          & 6.59           & 20.62          & 15.56          & 72.02           \\
T-CVAE \cite{Wang19TCVAE}         & 14.34          & 5.06          & 2.01          & 1.13          & 4.23           & 15.51          & 11.49          & 53.77           \\
MG+Trans \cite{Huang21IgSEG}       & 19.43          & 7.47          & 3.92          & 2.46          & 7.63           & 19.62          & 14.42          & 74.95           \\
MG+CIA \cite{Huang21IgSEG}         & 20.91          & 7.46          & 3.88          & 2.35          & 7.29           & 21.12          & 19.88          & 82.89           \\
MGCL \cite{Huang21IgSEG}            & 22.57          & 8.16          & 4.23          & 2.49          & 7.84           & 21.66          & 21.46          & 88.41           \\
MMT \cite{Xue22MMT}            & 22.87          & 8.68          & 4.38          & 2.61          & 15.55          & 23.61          & 25.41          & 103.11          \\
MET (Ours)      & \textbf{24.31} & \textbf{8.79} & \textbf{4.62} & \textbf{2.73} & \textbf{16.41} & \textbf{24.49} & \textbf{26.47} & \textbf{107.82} \\ \bottomrule
\end{tabular}
\caption{Comparison results on VIST-E test set. B@n, M, R-L, C and rSUM denote BLEU@n, METEOR, ROUGE-L, CIDEr and Result Sum, respectively.}
\label{tab:exp}
\end{table*}

\begin{table*}[t]
\centering
\begin{tabular}{lcccccccc}
\toprule
\textbf{Method} & \textbf{B@1}   & \textbf{B@2}  & \textbf{B@3}  & \textbf{B@4}  & \textbf{M}     & \textbf{R-L}   & \textbf{C}     & \textbf{rSUM}  \\\midrule
Seq2Seq \cite{Luong15Effective}         & 14.21          & 4.56          & 1.70          & 0.70          & 11.01          & 19.69          & 8.69           & 60.56          \\
Transformer \cite{Vaswani17Attention}    & 15.35          & 4.49          & 1.82          & 0.76          & 11.43          & 19.16          & 9.32           & 62.33          \\
MGCL \cite{Huang21IgSEG}           & 15.89          & 4.76          & 1.57          & 0.00          & 11.61          & 20.30          & 9.16           & 63.29          \\
MMT \cite{Xue22MMT}            & 18.52          & 5.99          & 2.51          & 1.13          & 12.87          & 20.99          & 12.41          & 74.42          \\
MET (Ours)      & \textbf{19.98} & \textbf{6.48} & \textbf{2.89} & \textbf{1.77} & \textbf{14.53} & \textbf{22.73} & \textbf{13.85} & \textbf{82.23} \\ \bottomrule
\end{tabular}
\caption{Comparison results on LSMDC-E test set.}
\label{tab:exp2}
\end{table*}

\subsection{Implementation Details}
For the scene graph, we limit the maximum number of objects to 10 and the maximum number of relationships to 20. The relational graph convolution network includes four relational graph convolution layers, and the size of input and output sets of 768. For semantic event reasoning, we use a pre-trained BERT model \cite{Devlin19BERT} as the language model. The layers and attention heads of the decoder are 12 and 8. The dimension of embedding vectors in the decoder is 768, and the dimension of hidden states is 768. The visual feature encoder is ResNet-152. For model training, we select the Adam optimizer \cite{KingmaB14} to optimize the model with learning rate of 2e-4. The maximum training epoch of our model is 25. The trade-off parameter $\alpha$ in Eq.\ref{equ:loss} is 0.2. The batch size, weight decay and warm-up proportion are 128, 0.01 and 0.1. During inference, we use the beam search with a beam size of 3 to generate a story ending with maximum sentence length is 25. Our model is trained on one V100 GPU.
\begin{table*}[t]
\centering
\begin{tabular}{lcccccccc}
  \toprule
\textbf{Method} & \textbf{B@1}   & \textbf{B@2}  & \textbf{B@3}  & \textbf{B@4}  & \textbf{M}     & \textbf{R-L}   & \textbf{C}     & \textbf{rSUM}   \\\midrule
MET      & \textbf{24.31} & \textbf{8.79} & \textbf{4.62} & \textbf{2.73} & \textbf{16.41} & \textbf{24.49} & \textbf{26.47} & \textbf{107.82} \\
w/o ID          & 23.84          & 8.70          & 4.51          & 2.56          & 15.91          & 24.10          & 25.86          & 105.48          \\
w/o CMF         & 23.47          & 8.65          & 4.47          & 2.53          & 15.91          & 23.85          & 25.66          & 104.54          \\
w/o MI          & 22.68          & 8.56          & 4.33          & 2.48          & 15.83          & 22.99          & 24.74          & 101.61          \\
w/o VER         & 22.41          & 8.25          & 4.33          & 2.50          & 15.86          & 23.09          & 25.03          & 101.47          \\
w/o SER         & 23.78          & 8.73          & 4.46          & 2.55          & 15.88          & 24.04          & 25.87          & 105.31          \\
w/o CMF, MI     & 21.03          & 8.03          & 4.16          & 2.36          & 15.43          & 21.14          & 22.44          & 94.59         \\\bottomrule    
  \end{tabular}
  \caption{Ablation study. 
    ``\textit{w/o ID}'' denotes removing the incoherence detection objective;
    ``\textit{w/o CMF}'' denotes removing the cross-modal fusion; 
    ``\textit{w/o MI}'' denotes removing the multimodal injector; 
    ``\textit{w/o VER}'' denotes removing the event-based reasoning in visual event graph; 
    ``\textit{w/o SER}'' denotes removing the event-based reasoning in semantic event graph; 
    ``\textit{w/o CMF, MI}'' removing the cross-modal fusion and multimodal injector.}
  \label{tab:abl}
\end{table*}

\subsection{Baselines}
We compare our model with following state-of-the-art baselines: (1) {\bf Seq2Seq} is a stack RNN-based model \cite{Luong15Effective} with attention mechanisms, and image features are directly concatenated. (2) {\bf Transformer}, proposed by \citet{Vaswani17Attention}, is an encoder-decoder model with self-attention mechanisms. (3) {\bf IE+MSA} is a story ending generation model incorporating external knowledge \cite{Guan19Story}. (4) {\bf T-CVAE} \cite{Wang19TCVAE} is a conditional variational autoencoder based on transformer for missing story plots generation. (5) {\bf MG+Trans} consists of multi-layer graph convolutional networks and a transformer decoder \cite{Huang21IgSEG}. (6) {\bf MG+CIA} consists of multi-layer graph convolutional networks, top-down LSTM and one context-image attention unit in the decoder \cite{Huang21IgSEG}. (7) {\bf MGCL} is an image-guided story ending generation model with multi-layer graph convolution networks and cascade-LSTM \cite{Huang21IgSEG}. (8) {\bf MMT} is a multimodal memory transformer for image-guided story ending generation \cite{Xue22MMT}.

\subsection{Main Results}
The experimental results on VIST-E and LSMDC-E are shown in Table \ref{tab:exp} and Table \ref{tab:exp2}. From the tables, we can make two observations. Firstly, our model achieves state-of-the-art performance on the VIST-E and LSMDC-E datasets compared to other strong competitors. In addition, MG+CIA, MGCL, MMT and our model significantly and consistently outperform other models that directly concatenate visual features. It indicates that mining visual information is essential and can provide rich information to predict the ending. Moreover, our model achieves better results than MG+CIA, MGCL and MMT, demonstrating that reasoning and mining implicit information from story plots and ending image is significant for image-guided story ending generation.

\subsection{Ablation Study}
To verify the effectiveness of our method, we conduct an ablation study and show the results in Table \ref{tab:abl}. Firstly, the table shows that removing each component or objective decreases the model performance, which demonstrates our method's effectiveness. In addition, we observe that removing cross-modal fusion and multimodal injector brings a great performance drop, which shows that cross-modal information mining and adaptive integration play a crucial role in story ending prediction.

\subsection{SEG Setting}
\begin{table}[t] 
  \centering
  \setlength{\tabcolsep}{4.3pt}{
  \begin{tabular}{lccccc}
  \toprule
  \textbf{Method} & \textbf{B@1}   & \textbf{B@2}  & \textbf{B@4}  & \textbf{M}    & \textbf{R-L}   \\\midrule
  Seq2Seq         & 14.27          & 4.27          & 1.05          & 6.02          & 16.32          \\
  Transformer     & 17.06          & 6.18          & 1.57          & 6.55          & 18.69          \\
  IE+MSA          & 20.11          & 6.62          & 1.68          & 6.87          & {\ul 21.27}    \\
  T-CVAE          & 20.36          & 6.63          & 1.88          & 6.74          & 20.98          \\
  Plan\&Write      & 20.92          & 5.88          & 1.44          & 7.10          & 20.17          \\
  KE-GPT2         & \textbf{21.92} & \textbf{7.40} & 1.90          & {\ul 7.41}   & 20.58          \\
  MG+Trans        & 18.55          & 6.76          & {\ul 2.33}    & 7.31          & 19.02          \\
  MGCL            & 20.27          & 6.26          & 1.81          & 6.91          & 21.01          \\
  MET           & {\ul 21.88}    & {\ul 7.28}    & \textbf{2.36} & \textbf{7.41}    & \textbf{21.32}     \\\bottomrule
  \end{tabular}}
  \caption{Result of the SEG task on the VIST-E dataset (plain text). The bold / underline denotes the best and the second performance, respectively.}
  \label{tab:seg}
\end{table}
To investigate the effectiveness of visual information mining in our method, we remove the image from the VIST-E dataset and evaluate it on only plain text. The results are shown in Table \ref{tab:seg}. From the table, we observe that our model keeps competitive with Plan\&Write \cite{Yao19Plan} and KE-GPT2 \cite{Guan20Knowledge} models designed especially for textual story generation. Moreover, our model outperforms MG+trans, which verifies the effectiveness of our incoherence detection and semantic event-based reasoning. Our model performs better when adding the image, as shown in Table \ref{tab:exp}. It demonstrates that mining implied visual information can help story ending generation.

\subsection{Analysis}
\subsubsection{Impact of Event-based Reasoning}
\begin{table*}[t]
\centering
\begin{tabular}{lcccccccc}
\toprule
\textbf{Method}        & \textbf{B@1}   & \textbf{B@2}  & \textbf{B@3}  & \textbf{B@4}  & \textbf{M}     & \textbf{R-L}   & \textbf{C}     & \textbf{rSUM}   \\ \midrule
MET                    & \textbf{24.31} & \textbf{8.79} & \textbf{4.62} & \textbf{2.73} & \textbf{16.41} & \textbf{24.49} & \textbf{26.47} & \textbf{107.82} \\
w/ Dependence Parser   & 23.47          & 8.70          & 4.50          & 2.57          & 15.88          & 24.15          & 24.06          & 103.33          \\
w/o Visual Event Graph & 22.13          & 8.19          & 4.21          & 2.44          & 15.76          & 22.88          & 23.93          & 99.54           \\
w/o CMF                & 23.47          & 8.65          & 4.47          & 2.53          & 15.91          & 23.85          & 25.66          & 104.54     \\\bottomrule    
  \end{tabular}
  \caption{Impact of event reasoning.
  ``\textit{w/ Dependency Parser}'' denotes replacing semantic role labeling with dependency parsing;
  ``\textit{w/o Visual Event Graph}'' denotes removing the visual event graph and provides the whole image features as inputs; 
  ``\textit{w/o CMF}'' denotes removing the cross-modal fusion.}
  \label{tab:imp}
\end{table*}
To investigate the effectiveness of event reasoning, we analyze its impact, and the results are shown in Table \ref{tab:imp}. From the table, we can observe that replacing semantic role labeling with dependency parsing leads to decreased performance. Moreover, replacing the visual event graph with whole image features (i.e., features extracted by pre-trained CNN) shows a performance drop. In addition, removing cross-modal fusion also shows a performance drop. These demonstrate the effectiveness of event-based reasoning for the image-guided story ending generation.

\subsubsection{Case Study}
\begin{figure*}[t]
  \centering
  \includegraphics[width=\linewidth]{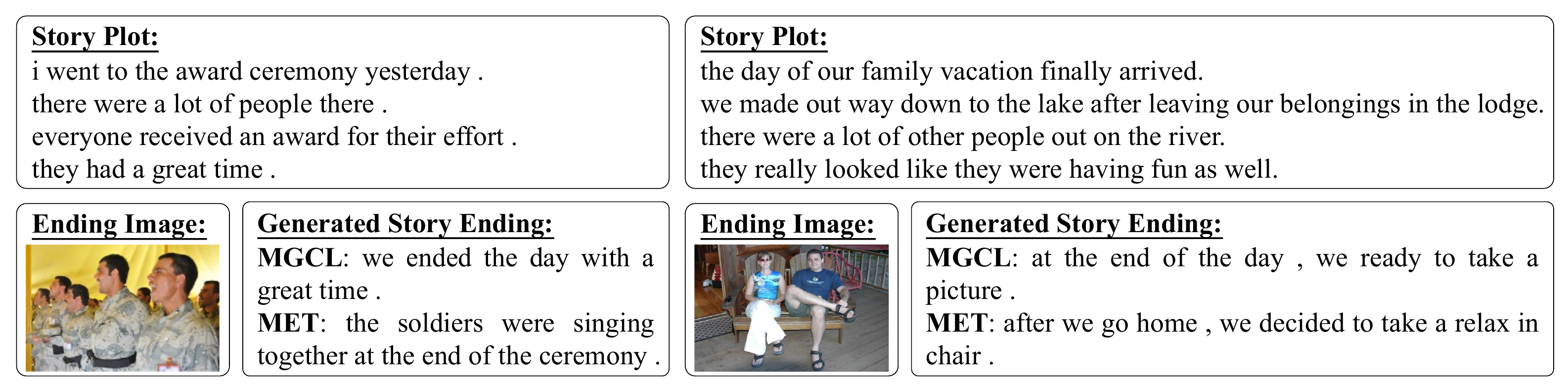}
  \caption{Random sampling examples generated by MET and MGCL.}
  \label{fig:case}
\end{figure*}
To extensively evaluate our method, we conduct a case study for our model and MGCL, and some random sampling examples are shown in Figure~\ref{fig:case}. For example, in the left case, we can observe that our model can reason that the man in the image is a soldier, while the result from MGCL is not significantly related to visual content. For example, in the right case, our model can generate the word "relax" based on the objects "human" and "chair". It shows that our model can mine the implicit information based on visual and semantic information.

\subsubsection{Interpretable Visualization Analysis}
\begin{figure}[t]
  \centering
  \includegraphics[width=\linewidth]{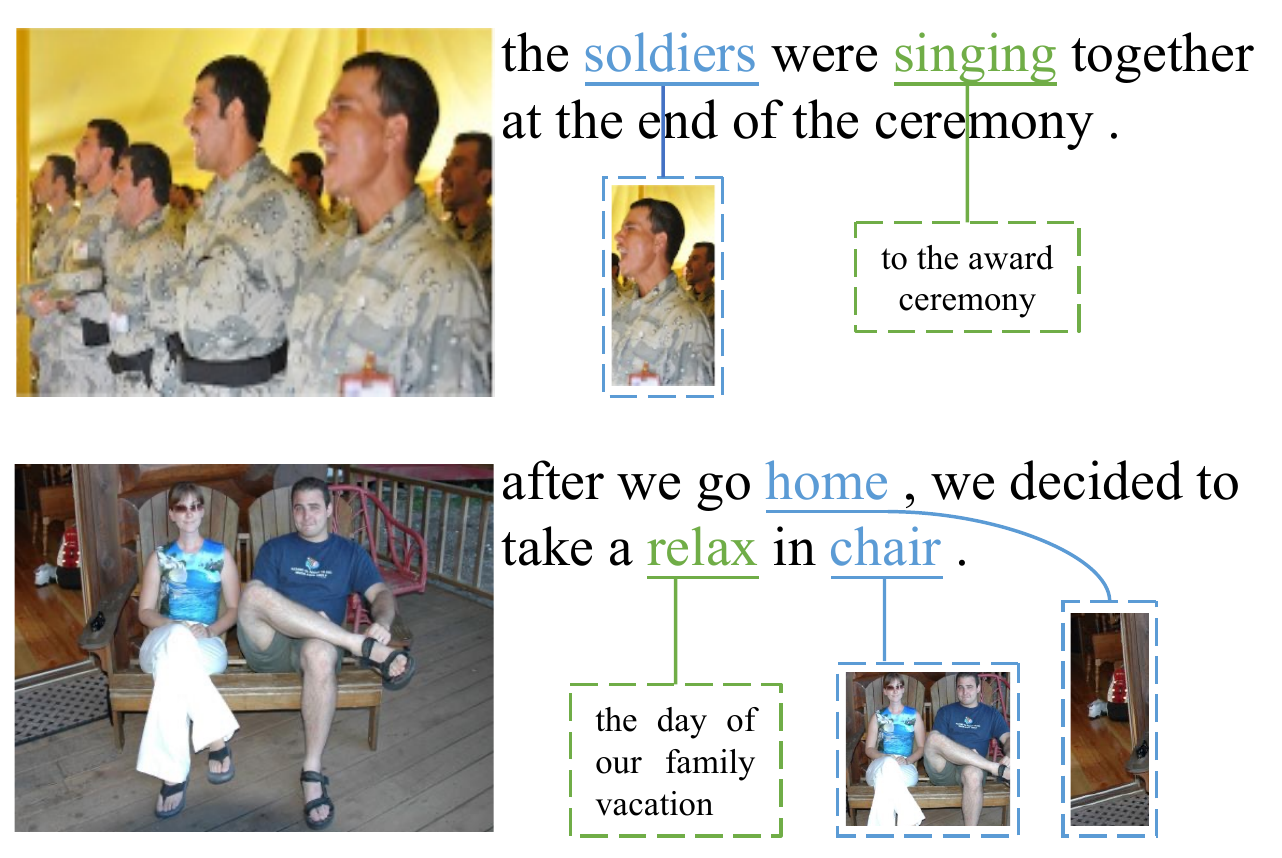}
  \caption{Interpretable visualization analysis of our method (better viewed in color).}
  \label{fig:visual}
\end{figure}
\begin{table}[t]
  \centering
  \begin{tabular}{lccc}
  \toprule
  \textbf{Method} & \textbf{Gram.} & \textbf{Logic.} & \textbf{Rele.} \\\midrule
  MET           & \bf 3.49           & \bf 3.37            & \bf 2.94           \\
  MGCL            & 3.36           & 3.15            & 2.66           \\
  MG+Trans        & 3.22           & 2.78            & 2.71           \\\bottomrule
  \end{tabular}
  \caption{Human evaluation.}
  \label{tab:hum}
\end{table}
To investigate the effectiveness of the multimodal injector, we conduct an interpretable visualization analysis. The results are shown in Figure \ref{fig:visual}. The word with a blue underline denotes that the multimodal injector is assigned the greater probability in the node of visual event graph. Green corresponds to greater probability in the node of semantic event graph. The dotted boxes below represent the specific content of nodes. From the results, we can observe that nodes in visual and semantic event graphs are able to deduce implicit information.

\subsubsection{Human Evaluation}
To evaluate our method more comprehensively, we conducted a human evaluation to compare further the performance of our model and MGCL and MG+trans. As follow \citet{Huang21IgSEG}, we considered three metrics for the story ending generated by models: Grammaticality (Gram.) \cite{Wang19TCVAE} evaluates correctness, natural, and fluency of story endings; Logicality (Logic.) \cite{Wang19TCVAE} evaluates reasonability and coherence of story endings; Relevance (Rele.) \cite{Yang19Knowledgeable} measures how relevant between images and generated story endings. We randomly sampled 100 samples from the test set and display them to 3 recruited annotators. Thereby, each annotator worked on 300 items from 3 models. We show 3 annotators all outputs from all 3 models at once and shuffle the output-model correspondence to ensure that annotators do not know which model the output is predicted from. Following \citet{Yang19Knowledgeable}, we set a 5-grade marking system, where one is the worst grade, and five is the maximum. The results show that the performance of our model is significantly better than MGCL and MG+trans. That is, our model can generate higher-quality story endings.

\section{Conclusion}
In this work, we propose a multimodal event transformer, a framework for image-guided story ending generation. Our method includes event graph construction, event-based reasoning, cross-model fusion, multimodal injector and story ending generation. Different from previous work, our method not only focuses on cross-modal information fusion but also on reasoning and mining implicit information from single-modality data. In addition, we propose an incoherence detection to enhance the understanding context of a story plot and robustness of graph modeling for our model. In the experiments, results show that our method delivers state-of-the-art performance. 

\section*{Limitations}
Although our proposed method can effectively reason and mine implicit information from story plots and ending image, it suffers from weaknesses in integrating cross-modal information. Specifically, our method connects visual and semantic event graphs by connecting whole image nodes and whole sentence nodes. It lacks fine-grained information to pass between semantic events to visual objects. In further work, we will study how to pass fine-grained information between visual and semantic event graphs.

\bibliography{ref}
\bibliographystyle{acl_natbib}

\end{document}

%% file: math_commands.tex
\usepackage{amsmath,amsfonts,bm}

\def\eqref#1{equation~\ref{#1}}
\def\1{\bm{1}}

\def\vh{{\bm{h}}}

\def\vp{{\bm{p}}}

\def\vr{{\bm{r}}}
\def\vs{{\bm{s}}}
\def\vt{{\bm{t}}}

\def\vv{{\bm{v}}}
\def\vw{{\bm{w}}}

\def\mC{{\bm{C}}}

\def\mI{{\bm{I}}}

\def\mS{{\bm{S}}}
\def\mT{{\bm{T}}}
\def\mU{{\bm{U}}}
\def\mV{{\bm{V}}}
\def\mW{{\bm{W}}}

\DeclareMathAlphabet{\mathsfit}{\encodingdefault}{\sfdefault}{m}{sl}
\SetMathAlphabet{\mathsfit}{bold}{\encodingdefault}{\sfdefault}{bx}{n}

\def\gE{{\mathcal{E}}}

\def\gG{{\mathcal{G}}}

\def\gL{{\mathcal{L}}}

\def\gN{{\mathcal{N}}}
\def\gO{{\mathcal{O}}}

\def\gR{{\mathcal{R}}}
\def\gS{{\mathcal{S}}}

\def\gV{{\mathcal{V}}}

\newcommand{\R}{\mathbb{R}}

\newcommand{\softmax}{\mathrm{softmax}}
\newcommand{\sigmoid}{\sigma}